\documentclass[conference]{IEEEtran}
\IEEEoverridecommandlockouts
\usepackage{cite}
\usepackage{amsmath,amssymb,amsfonts}
\usepackage{algorithmic}
\usepackage{graphicx}
\usepackage{textcomp}
\usepackage{xcolor}
\def\BibTeX{{\rm B\kern-.05em{\sc i\kern-.025em b}\kern-.08em
    T\kern-.1667em\lower.7ex\hbox{E}\kern-.125emX}}
        
\usepackage{fancyhdr}
\begin{document}

\title{Detecting F-formations \& Roles in Crowded Social Scenes with Wearables: Combining Proxemics \& Dynamics using LSTMs\\
\thanks{ This paper was partially funded by the
Dutch national program COMMIT and the Netherlands Organization for Scientific Research (NWO) under project number 639.022.606}
}

\author{\IEEEauthorblockN{Alessio Rosatelli}
\IEEEauthorblockA{\textit{Ingegneria Informatica e Robotica} \\
\textit{University of Perugia}\\
Perugia, Italy \\
alessio.rosatelli1994@gmail.com}
\and
\IEEEauthorblockN{Ekin Gedik}
\IEEEauthorblockA{\textit{Socially Perceptive Computing Lab} \\
\textit{Delft University of Technology}\\
Delft, Netherlands \\
e.gedik@tudelft.nl}
\and
\IEEEauthorblockN{Hayley Hung}
\IEEEauthorblockA{\textit{Socially Perceptive Computing Lab} \\
\textit{Delft University of Technology}\\
Delft, Netherlands \\
h.hung@tudelft.nl}

}

\maketitle
\thispagestyle{fancy}

\begin{abstract}
In this paper, we investigate the use of proxemics and dynamics for automatically identifying conversing groups, or so-called F-formations. More formally we aim to automatically identify whether wearable sensor data coming from 2 people is indicative of F-formation membership. We also explore the problem of jointly detecting membership and more descriptive information about the pair relating to the role they take in the conversation (i.e. speaker or listener). We jointly model the concepts of proxemics and dynamics using binary proximity and acceleration obtained through a single wearable sensor per person. We test our approaches on the publicly available MatchNMingle dataset which was collected during real-life mingling events. We find out that fusion of these two modalities performs significantly better than them independently, providing an AUC of 0.975 when data from 30-second windows are used. Furthermore, our investigation into roles detection shows that each role pair requires a different time resolution for accurate detection.
\end{abstract}

\begin{IEEEkeywords}
F-formation detection, wearable sensing, conversing groups, recurrent neural networks, role identification
\end{IEEEkeywords}

\section{Introduction}
As social animals, most people interact and specifically converse with each other on a daily basis. Regardless of the specific type of the event that the interaction takes place, the automatic detection of conversational interaction is the first step to understanding the potential flow of interpersonal influence. For example, being able to accurately identify conversing groups allows us to analyse more complex social phenomena such as attraction \cite{MadanetalHCI2005} or cohesion \cite{NanningaICMI2017}. The analysis of social behaviour is also shown to be crucial for security and surveillance applications \cite{garate2014group,Cristani2013HumanBA}. This information is beneficial for human robot interaction by allowing the robot to identify an ongoing conversation to enter \cite{RamirezVACC16}. In this paper, we investigate and provide a novel solution for the detection of conversing groups and participants roles in the interaction, relying solely on the fusion of multimodal sensor data, namely acceleration and proximity and test our solution on a real life crowded mingling event. 
\par
We formalise the notion of conversing groups by referring to Kendon's definition  F-formations \cite{kendon1990conducting}. Kendon defines F-formations as a specific type of focused encounter where participants tend to stay in close proximity and orient themselves to sustain conversation \cite{kendon1990conducting}. From the definition, it is clear that proxemics, the spatial distance and bodily orientation of participants, represents valuable information for the identification of F-formations. Indeed, most related work from the wearable and ubiquitous computing community heavily relies on proxemics, extracting proximity and orientation information either from images \cite{setti2015f,bazzani2013social,vascon2016detecting} and/or wearable sensing such as Infrared trans-receivers \cite{alameda2015analyzing,gips2006alex}. However, there are limitations to such approaches. The estimation of spatial proximity with IR or radio-based sensors are shown to result in various false detections, mainly caused by the surrounding environment and furniture \cite{chaffin2017promise}. In crowded scenarios, possible occlusions between people and objects can negatively affect the robust estimation of proximity and orientation. In this paper, we propose an alternate approach fusing the proxemics and dynamics of interaction, based on the findings that people tend to coordinate their movements during conversational interaction \cite{kendon1970movement}. This builds upon the work of Gedik and Hung \cite{gedik2018detecting} where they proposed an approach that investigate dynamics alone. Here we argue for the importance of the use of both for better accuracy.

\par
We investigate the fusion of proxemics and dynamics with a publicly available dataset of real-life crowded mingling events \cite{cabrera2018matchnmingle}. Such events are perfect candidates for analysing group behaviour since groups of different sizes form and break over time naturally, creating rich and varied behaviour to analyse. During the mingling events used in our experiments, participants were fitted with a single sensor pack that was worn over their necks with a lanyard. These sensor packs recorded their bodily acceleration and the IDs of neighbouring sensor packs, acting as binary proximity detections. Our proposed solution uses these information sources as inputs to an Long Short Term Memory (LSTM) network which models the temporal dependencies of the data and learns a joint representation of proxemics and dynamics. Subsequently, we delve deeper into the nature of the problem by investigating the detection of the roles of the participants in the interaction in terms of whether they are speaking or listening. Based on findings in social science, speakers and listeners behave differently in interaction \cite{kendon1970movement}. We expect these roles to be identifiable and that they could eventually help in distinguishing conversing partners who are more or less involved with each other. 

\par

The novel contributions of this paper are as follows: (i) we propose to fuse proxemics and dynamics through LSTM networks and show, with a specific time resolution, it significantly outperforms using separate modalities, (ii) we investigate the automatic detection of roles inside the interaction and find that:
\begin{itemize}
\item Group membership and roles in interaction can be jointly detected.
\item Each role pair requires a different temporal resolution of analysis for satisfactory detection.
\end{itemize}

\section{Related Work}
To better contextualise the nature of this paper, we present a number of studies that focus on the automatic identification of F-formations. We use a rough categorisation based on the sensor types used in these studies. Most of the related literature in this domain focuses on the use of proxemics, the positions and orientations of the participants in a scene, as a cue. Classically favoured modalities for F-formation detection has been static images and videos. Throughout the years, many different methodologies such as Hough voting \cite{BMVC.25.23,setti2013multi}, graph clustering with dominant sets \cite{hung2011detecting,gan2013temporal} and multi-payoff evolutionary game theory \cite{vascon2016detecting} were used to detect F-formations from static images or video. 

\par

Another widely used sources of information are wearable and mobile sensing. Such devices provide proximity information through Infrared (IR) and/or radio which is then used to detect interactions between people. In general, the sensor data is taken at face value and used directly as a proxy for conversational interaction which can be used to infer F-formations. However, most works mentioned here do not evaluate the robustness of such sensing systems since most studies aggregate observed face to face detections over sufficiently long period of time with the assumption of relatively low crowd density. Such studies have tended to use these interaction proxies for analysing long term social concepts such as centrality \cite{choudhury2003sensing}, personality traits \cite{olguin2009sensible}, social patterns in daily life \cite{eagle2006reality}, student mental health \cite{wang2014studentlife}, interest and affiliation \cite{gips2006alex} and dynamics of interaction networks \cite{cattuto2010dynamics}. These studies might use custom-made sensor packs \cite{choudhury2003sensing,gips2006alex,cattuto2010dynamics} or employ mobile phones \cite{eagle2006reality,wang2014studentlife}. The temporal resolution of the analysis can range from minutes \cite{cattuto2010dynamics,gips2006alex} to days \cite{choudhury2003sensing} to months \cite{wang2014studentlife,eagle2006reality}. Some studies explicitly infer body orientation in addition to proximity \cite{matic2012analysis}, whereas some others infer the proximity from GPS coordinates \cite{casagranda2015audio}. In \cite{alameda2015analyzing}, information from wearable sensors is employed together with information from images for F-formation detection. 

\par

There exist few studies that consider the dynamics of interaction as cues for F-formation detection. Authors proposed two methods for modelling the dynamics of proxemics from video data in \cite{RamirezVACC16}. In \cite{choudhury2003sensing}, audio is used to infer speaking status which was then used for refining the IR results on interaction detection. Similarly, \cite{hung2014detecting} used acceleration to infer various social actions such as speaking, drinking, etc. Mutual information between different participants' social action streams were then computed and thresholded to obtain F-formation memberships. \cite{gedik2018detecting} built on these results and proposed new measures between social action streams and group size based training. To our knowledge, no other studies on F-formation detection focused on fusing the proxemics and dynamics of interaction from wearables.

\par
Determining similar studies for role identification, in the context of this paper, is not trivial. Identification of speakers mainly falls under speaker detection \cite{brummer2006application} or recognition \cite{campbell1997speaker}. They aim to find out who the speaker is from an input stream. These studies mostly employ sensors that registers the flow of speech, such as audio \cite{brummer2006application,campbell1997speaker} or video that focuses on the movement of the mouth \cite{cutler2000look}. Only a few studies employ accelerometers for detecting speakers, similar to our proposed approach \cite{gedik2017personalised}. However, none considers the roles in the context of an interaction and focuses on the joint estimation as our paper does. The benefit of modelling both jointly is that the correlation between the different behaviours should help us to better estimate conversational involvement and therefore social influence from the perspective of speaker and listener behaviour.

\section{Data}
In our experiments, we used the data from the MatchNMingle dataset which includes data from three separate speed dating events took place in a local bar and joined by 92 participants in total \cite{cabrera2018matchnmingle}. In each event, speed dating sessions were followed by a free mingling session of roughly one hour. In the mingling session, participants freely interacted with each other. Each participant wore a single custom badge-like sensor pack on their necks with a lanyard. These sensor packs contain an accelerometer and a radio-based binary proximity sensor, recording at 20Hz and 1Hz, respectively. The recordings of these sensor packs during the mingling sessions are used in our experiments. The video recordings of the scene were only used for obtaining F-formation and speaking status annotations which has been used as ground truth in the roles detection part of this study. 10 minute intervals from the mingling session of each day, totalling in 30 minutes, were annotated for F-formations. Sensor readings and annotations from these 30 minutes form the data used in this paper. During these 30 minutes, groups of various sizes ranging from dyadic interactions to seven people were formed, creating a rich testing environment. For more details regarding the dataset, please refer to \cite{cabrera2018matchnmingle}.

\begin{figure*}[h!]
        \centering
        \includegraphics[width=0.9\textwidth]{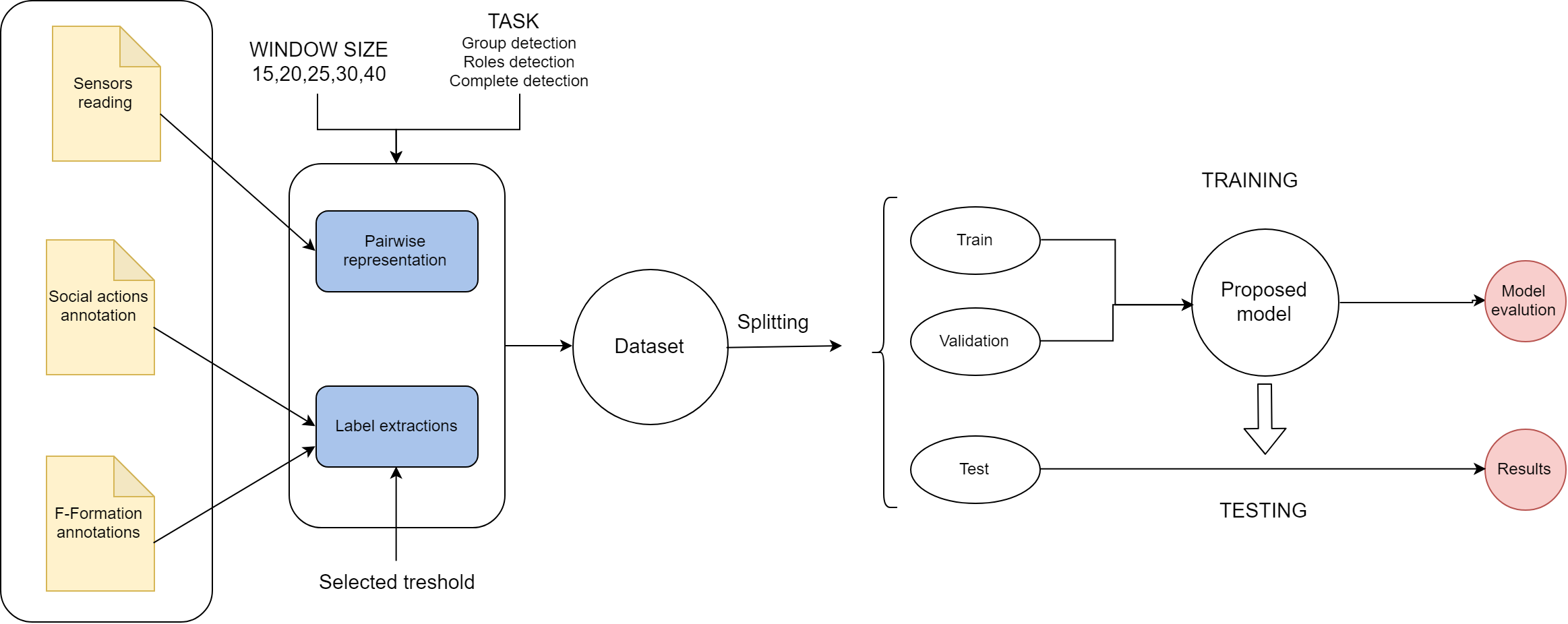}
        \caption{Flow diagram of the proposed method}
        \label{fig:workflow}
    \end{figure*}
\section{Methodology}

Figure \ref{fig:workflow} shows an overview of the proposed method and conducted experiments. In this section, the pairwise representation procedure which creates the samples used in our experiments will be first explained. Then, the methodologies and experiment setups for group membership and roles detection will be presented. 

\subsection{Pairwise representation}
    \begin{figure}
    \centering
        \includegraphics[width=0.7\linewidth]{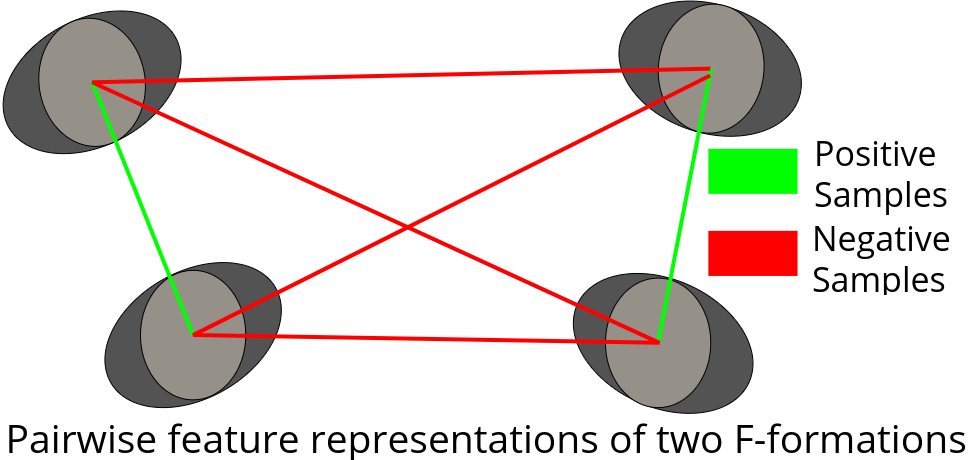}
        \caption{Synthetic visualisation of pairwise representations (taken from \cite{gedik2018detecting}) }
        \label{fig:2_f_form}
    \end{figure}
    According to the literature, a group could be represented as a collection of individuals \cite{bazzani2013social} or as a set of dyadic pairs \cite{hung2011detecting,gedik2018detecting,inaba2016conversational}. In this paper, we selected pairwise representations since we utilise, in addition to proxemics, the dynamics of interaction realised through the coordination of the movements of participants. Figure \ref{fig:2_f_form} shows a synthetic visualisation of a possible scene with four people and two F-formations. Pairwise representations of such a scene treat each possible connection between people as a sample. These samples are visualised in Figure \ref{fig:2_f_form} as green and red lines, representing samples belonging to the positive (pair is in the same F-formation) and the negative (pair is not in the same F-formation) classes, respectively.   
    \par
    Since we are interested in the dynamics of interaction which unfolds over time, the temporal resolution of the data gains importance \cite{gan2013temporal,vascon2016detecting}. An instantaneous representation of the scene is not sufficient for exploiting the coordination patterns between interacting partners. Hence, we used a sliding window approach, with 50\% overlap, over the sensor data of participants. Samples in our experiments are formed by continuous sequences of sensor data obtained from two participants. Following sections will provide more information about the dimensions of these input samples with respect to experiment setups and chosen sliding window sizes.
    
    \subsection{Group membership detection}
            \begin{figure*}
            \centering
            \includegraphics[width=0.85\textwidth]{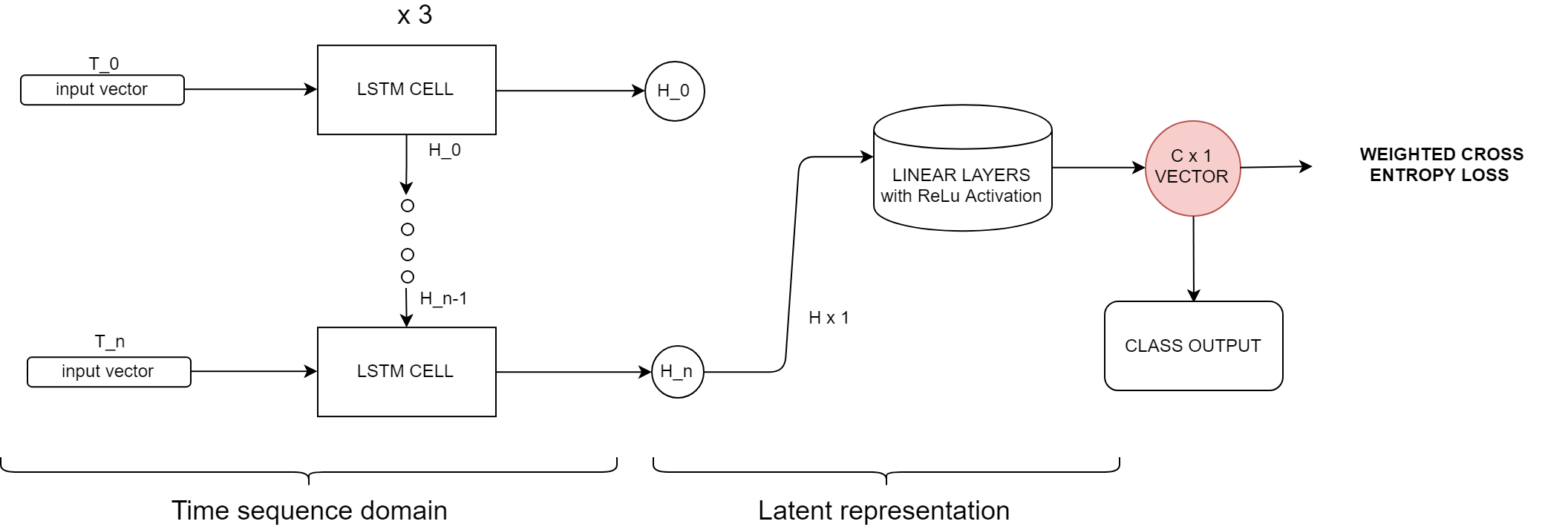}
            
            \caption{Proposed model structure and information propagation through the network.}
            \label{fig:architecture}
        \end{figure*}
  
    Since our samples are represented as time-series data, we used recurrent neural networks. The proposed architecture is visualised in Figure \ref{fig:architecture}. Each sample is fed into a three-layer LSTM \cite{hochreiter1997long} with 16 hidden units per cells. The cell number varies according to the selected window size and equals to the input sequence length. The number of layers and hidden units were empirically found. The outputs of the last LSTM layer cells are discarded except the last one which should ideally contain the information extracted from the previous elements of the sequences. This representation vector is then fed into a feed-forward network which is composed of two linear layers with Relu activation functions. This network reduces the vector dimension from 16 to 8 and then maps it into an output vector with dimensions identical to the number of classes for the current experiment.
    \par
    The model is trained end to end with a weighted cross entropy loss on the output vector, defined by the equation below:

    $loss(x, class) = weight[class] \left(-x[class] + \log\left(\sum_j \exp(x[j])\right)\right)$

    The pairwise representation creates high class imbalance in the data, favouring the negative samples. Hence, we used a weighted loss where the weights were set to be inversely proportional to the label distribution. The network was optimised with Adam. \cite{kingma2014adam}.
    \begin{table}
    \caption{Available input sources with respective IDs (P1 and P2 stand for Participants 1 and 2)}
    \begin{center}
        \begin{tabular}{|l|l|} 
        \hline
        Input ID & Definition                                 \\ 
        \hline
        0 : 2      & X,Y,Z accelerometer readings for P1  \\ 
        \hline
        3 : 5      & X,Y,Z accelerometer readings for P2  \\ 
        \hline
        6          & Binary proximity sequence between P1 and P2 \\
        \hline
        \end{tabular}
        \label{tab: sample_features}
    \end{center}
    \end{table}

    As shown in Figure \ref{fig:2_f_form}, there are two classes for group detection experiments; pairs that are in the same F-formation and pairs that are not. A sample is treated as positive if the two participants forming the sample are annotated as being in the same group in the ground truth at least 66\% of the sliding window that the sample is extracted from. 
    
    We experiment with three different input combinations: acceleration only, proximity only and fusion of the two. Figure \ref{tab: sample_features} shows the inputs used in these combinations. For the first setup, tri-axial accelerations from two participants are concatenated to form the samples (IDs 0 to 5). The second setup uses the binary proximity stream only (ID 6). Samples for the fusion setup is formed by the concatenation of the acceleration and the proximity streams (IDs 0 to 6). In order to have the same sampling rate, binary proximity readings are up-sampled to 20Hz. As an example, with a window size of 15 seconds and an input combination of fusion, each sample will be 300x7. The final size of the resulting dataset varies according to the window size, both in the number of samples, which is conversely proportional to the window size, and the sequence length of each sample which is directly proportional to it. 
    
    \subsection{Joint prediction of membership and roles}
    We define two basic roles in an interaction: speaker and listener. Since we use a pairwise representation, this results in three different classes for each sample from pairs that were already in an interaction: speaker\&speaker, speaker\&listener and listener\&listener. We use the speaking status ground truth present in the MatchNMingle \cite{cabrera2018matchnmingle} to label someone as a speaker or a listener. If a person is annotated as speaking at least 30\% of the window that the sample is extracted from, this person is labelled as a speaker and listener otherwise. This threshold was selected to allow a continuous turn-taking behaviour to properly represent speaker\&speaker pairs.
    \par
    We formulate the experiments on role detection as four-class classification problems that focus on jointly detecting the existence of an interaction and the roles participants take in it. These joint prediction experiments include all samples used in the group detection experiments but the positive samples are now relabelled with respect to the role pairs. This formulation results in the following classes: pairs that are not in the same F-formation, speaker\&speaker pairs, speaker\&listener pairs and listener\&listener pairs.  

\section{Results and Discussion}

Following subsections present the results of the group and roles detection experiments. For both experiment types, we used 80-10-10 training-validation-test split. While splitting, we made sure there are no samples belonging to the same pair of people in the train, validation and test sets to avoid contamination. 

To test the generalisation ability of our methods, we repeated the random splitting step 20 times, resulting in 20 different networks and 20 performances for each. Results for group membership detection are then presented as the means of these 20 repetitions. For the joint role and membership detection experiments, confusion matrices for each run are accumulated and normalised at the end, for a more detailed evaluation in the multiclass case. Each model is trained for 50 epochs on the training set and the model with the lowest loss on the validation set is used for evaluation on the test set. 

\subsection{Group membership detection}

    \begin{figure}
        \includegraphics[width=0.88\linewidth]{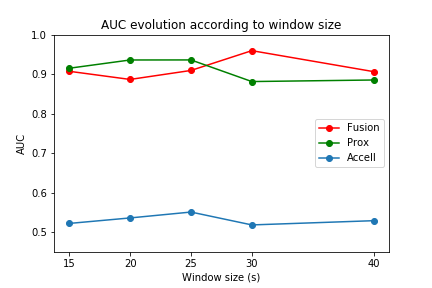}
        \caption{Mean AUC scores for group membership detection with different window lengths and inputs}
        \label{fig:Group membership result}
    \end{figure}
    
Due to highly imbalanced data, we selected Area Under Receiver Operating Characteristic Curve (AUC for short) as the evaluation metric. Figure \ref{fig:Group membership result} visualises the mean AUC scores obtained for group detection with different input combinations explained in the former section, with window sizes ranging from 15 to 40 seconds. Firstly, acceleration only results seem to be poor compared to the others, ranging from 0.52 to 0.57 (standard deviations are close to 0.06). Still, a one-sided paired t-test showed that the results are significantly better than a random classifier, regardless of the window size ($p<0.05$). We could also see that increasing the window size causes a marginal increase in the performance. This is expected since the dynamics of interaction, which we aim to capture through acceleration, are expected to unfold in larger time resolutions.
\par
We can see that proximity provides satisfactory performance by itself, providing AUC scores ranging from 0.87 to 0.94. Contrary to the pattern we saw for the acceleration only performances, performances with the proximity information tend to drop with the increasing window size. This is might suggest that proximity itself is not enough for detecting reasonably longer interactions. 
\par
Performances obtained with the fusion of acceleration and proximity supports our claims regarding the nature of interactions including both proxemics and dynamics. We see that the performances obtained with fusion tend to be slightly lower than the proximity ones for window sizes of 20 and 25. However, a paired one-tailed t-test showed no significant difference. In other words, fusion always guarantees performances at least as good as proximity. More interestingly, we can see that when 30-second windows are used, fusion provides a noticeable increase over the proximity with an AUC score of 0.975. This result is shown to be significantly better than the proximity one ($p<0.01$). When analysed together with the former results, we can conclude that dynamics captured through body acceleration indeed provides more information regarding the existence of an interaction between partners. However, it should be noted that this effect heavily depends on the size of the window, since the coordination of people can only be robustly sensed in specific time resolutions.   
\subsection{Joint prediction of group membership and roles}

       \begin{figure}
        \includegraphics[width=0.43\textwidth]{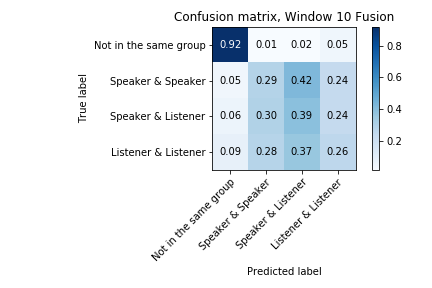}
        \caption{Joint prediction of group membership and roles confusion matrix for 10-second windows (accumulated over 20 repetitions and normalised)}
        \label{fig:CD_10}
        \end{figure}
    
\begin{figure*}[t!]
\centering
\begin{minipage}{.43\textwidth}
  \centering
  \includegraphics[width=\linewidth]{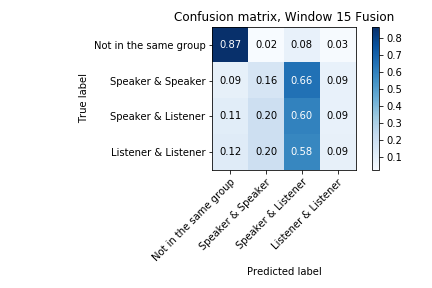}
  \caption{Group membership and roles confusion matrix for 15-second windows (accumulated over 20 repetitions and normalised)}
  \label{fig:CD_15}
\end{minipage}\hfill
\begin{minipage}{.43\textwidth}
  \centering
  \includegraphics[width=\linewidth]{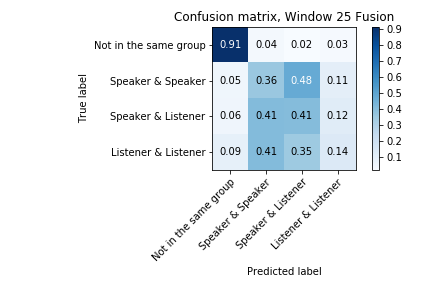}
  \caption{Group membership and roles confusion matrix for 25-second windows (accumulated over 20 repetitions and normalised)}
  \label{fig:CD_25}
\end{minipage}
\end{figure*}

Since we are dealing with a 4-class problem in joint prediction, we present normalised confusion matrices which make a more detailed evaluation possible. Following the group detection results, we stick to the fusion of proximity and acceleration for the experiments of the current section. Figures \ref{fig:CD_10}, \ref{fig:CD_15}, \ref{fig:CD_25} show the performances obtained with window sizes 10, 15 and 25, respectively. These three window sizes are empirically found out to be the best selections for the three distinct role pairs. For the 4-class classification, a random classifier is expected to obtain a normalised confusion matrix with scores of 0.25 in the diagonal, which acts as the baseline.

In all of the figures, we can see that our model easily identifies the no interaction class, providing normalised scores close to 0.9. This is compatible with the results of the former subsection where the proposed method with the fusion of both modalities successfully managed to distinguish between the presence and absence of an interaction between people. 

The proposed method is relatively successful in distinguishing between role pairs and the performances have some interesting implications. The highest performances for each role pair are obtained with different window sizes. For the speaker\&speaker class, the highest performance is obtained when 25 second windows are used. It is an expected result, since we already discussed that in order to detect coordination between interacting partners, longer time intervals are generally needed. This is especially true for speaker\&speaker pairs where the emergence of turn-taking patterns are necessary for robust identification. We also see that, regardless of the window size, many speaker\&speaker pairs are falsely labelled as speaker\&listener pairs. One possible cause for this is the speakers that were not expressive during interactions. Another one is the limited amount of samples for speaker\&speaker pairs in the training set where they were at least three times less than the others, resulting in them not being fully represented. On the other hand, results show that both other role pairs are equally misclassified as speaker\$speaker pairs. We can assume that listeners that were quite active in terms of their movement might be the reason for this. 

Highest performance for speaker\&listener pairs, 0.66, is obtained with the window size of 15 seconds. However, we also see that nearly two-thirds of other role pairs are also misclassified as speaker\&listeners. With this window size, no other role pair can be detected with a performance better than random. This suggests that the model frequently favoured the speaker\&listener pairs over others. This is not directly related to the distribution of the role pairs in the training since the percentages of speaker\&listener pairs were close to listener\&listener pairs. We can speculate that speaker\&listener pairs cover the largest variance of interaction patterns since they include two different roles, which might be the reason for these results.

For listener\&listener pairs, the only performance which is better than random is obtained with a window size of 10 seconds. The performance, 0.26, is only marginally higher than a random baseline. This is not surprising. There is no specific pattern for listener\&listener pairs as the turn-taking for speaker\&speaker pairs. Speaker\&listener pairs can be still detected through the coordination of the listeners' backchannels as they react to the speakers. Lack of such behavioural patterns for listener\&listener pairs make them relatively harder to model. These results show that more research into listener behaviour is needed if listener\&listener pairs need to be robustly detected.

These results present the first attempts on investigating a formerly understudied phenomenon. We can conclude that proxemics and dynamics provide valuable information on the roles of people in interaction. However, relatively low performances, especially for listener\&listener pairs, show that much work needs to be done for obtaining robust solutions. Next section discusses possible directions for further research.

\section{Conclusion and Future work }

In this study, we investigated the automatic detection of pairwise F-formation memberships and roles using wearable sensors. We built our solution on proxemics and dynamics of interaction which were sensed through proximity and acceleration. With an LSTM network, we were able to learn joint pairwise representations of these modalities. These representations are then used to identify the presence of interaction and the roles participant took in it. Our solution was fully automatic and trained end-to-end, converting raw input data to output labels. We tested our method on the publicly available MatchNMingle dataset which was collected during real-life mingling events. We also investigated different setups in terms of input modalities and time resolutions. We tested our method on 20 different splits of our dataset to evaluate its generalisation capabilities. 

Our experiments showed that using acceleration and proximity together guarantees performance at least as good as using separate modalities. With a window size of 30 seconds, which was sufficient for interaction dynamics to arise, the fusion of these modalities provided a mean AUC score of 0.975, significantly outperforming others. Our experiments on the detection of role pairs showed that different role pairs are better represented in different window sizes. Even though we were able to detect each role pair better than a random classifier, relatively low performances of these detections, especially for the listener\&listener pairs, showed there is still room for improvement.  
\par
We believe there are many possibilities for extending this work, both for group membership and role detection parts. First of all, our formulation of the problem was based on a pairwise representation of interactions. As a refinement step to group membership detection, pairwise predictions of our method can be used to create a complete scene representation, similar to a proximity graph. There exist studies in the literature which provide a solution for this step \cite{hung2011detecting}. By employing one, we can refine the predictions of our method. Secondly, our model treats the proximity and acceleration streams as equals while learning joint representations. More sophisticated formulations are also possible, such as using proximity streams as pooling layers for acceleration, which might result in better representation. 

The fixed threshold used to identify speakers for training might result in mislabelling some instances. For example, in a 5 person group, if everyone speaks for an equal amount of time in a given interval, their speaking lengths will correspond to one-fifth of the window. In our current setup, none of these people will be identified as speakers. A proper labelling approach which considers the group size should result in better identification. Moreover, previous work has already identified the importance of using the group size information in the training \cite{gedik2018detecting}. Using the group size as additional information in the training phase might cover more varied behaviours of different roles. Finally, in this work, we only focused on two basic roles: speakers and listeners. However, social science literature has vast knowledge of group behaviour and the roles people take in interaction. Identifying more representative classes and creating sophisticated models for representing them should result in more satisfactory results.
\bibliographystyle{IEEEtran}
\bibliography{sample}
\end{document}